\documentclass[10pt,twocolumn,letterpaper]{article}

\usepackage{iccv}
\usepackage{times}
\usepackage{epsfig}
\usepackage{graphicx}
\usepackage{amsmath}
\usepackage{amssymb}

\usepackage{algorithm}
\usepackage{algorithmic}
\usepackage{graphicx}
\usepackage{caption} 
\usepackage[section]{placeins}
\usepackage{verbatim}
\usepackage{bm}
\usepackage{subfigure}

\usepackage{enumitem}
\setitemize[1]{itemsep=-0.2em,topsep=0em}
\usepackage{graphicx}
\usepackage[font=small]{caption}
\usepackage{setspace}
\usepackage{multirow}
\usepackage{booktabs}
\usepackage{pifont}

\usepackage[pagebackref=true,breaklinks=true,letterpaper=true,colorlinks,bookmarks=false]{hyperref}

\iccvfinalcopy % *** Uncomment this line for the final submission

\def\iccvPaperID{1252} % *** Enter the ICCV Paper ID here

\ificcvfinal\fi
\begin{document}

%%%%%%%%% TITLE
\title{Knowledge Distillation via Route Constrained Optimization}

% \thanks{This work was done when Baoyun Peng was an intern at SenseTime Inc.}\hspace{4pt}

\author{
Xiao Jin$^{1}$\footnotemark[1]\hspace{4pt}, 
Baoyun Peng$^{2}$\thanks{Equal contribution.}\hspace{4pt}, 
Yichao Wu$^{1}$,
Yu Liu$^{3}$,
Jiaheng Liu$^{4}$,\\
Ding Liang$^{1}$, 
Junjie Yan$^{1}$, 
Xiaolin Hu$^{5}$\\
$^{1}$\hspace{2pt}SenseTime Research\hspace{50pt}
$^{2}$\hspace{2pt}National University of Defense Technology\hspace{2pt}\\
$^{3}$\hspace{2pt}The Chinese University of Hong Kong
\hspace{30pt}$^{4}$\hspace{2pt}Beihang University\hspace{30pt}
$^{5}$\hspace{2pt}Tsinghua University\\
}

\maketitle
%\thispagestyle{empty}

%%%%%%%%% ABSTRACT

\begin{abstract}
  Distillation-based learning boost the performance of the miniaturized neural network based on the hypothesis that the representation of a teacher model can be used as structured and relatively weak supervision, and thus would be easily learned by a miniaturized model. However, we find that the representation of a converged heavy model is still a strong constraint for training a small student model, which leads to a high lower bound of congruence loss. In this work, inspired by \cite{Bengio2009Curriculum} we consider the knowledge distillation from the perspective of curriculum learning by routing. Instead of supervising the student model with a converged teacher model, we supervised it with some anchor points selected from the route in parameter space that the teacher model passed by, as we called \textit{route constrained optimization (RCO)}. We experimentally demonstrate this simple operation greatly reduces the lower bound of congruence loss for knowledge distillation, hint and mimicking learning. On close-set classification tasks like CIFAR \cite{krizhevsky2009learning} and ImageNet \cite{deng2009imagenet}, RCO improves knowledge distillation by 2.14\% and 1.5\% respectively. 
  For the sake of evaluating the generalization, we also test RCO on the open-set face recognition task MegaFace.
  % Evaluation on the MegaFace \cite{kemelmacher2016megaface} open-set face recognition task shows that
   RCO achieves 84.3\% accuracy on 1 vs. 1 million task with only 0.8 M parameters, which push the SOTA by a large margin.
\end{abstract}

%%%%%%%%% BODY TEXT
\section{Introduction}
%------------------------------------------------------------------------
The performance of Convolutional Neural Network (CNN) can be significantly improved by the deeper and wider design of network structure.
Whereas, it is hard to deploy these heavy networks on energetic consumption processor with limited memory. One way to deal with this situation is to make a trade-off between performance and speed by designing a miniaturized model to reduce the computational workload. Thus, narrowing the performance gap between heavy model and miniaturized model becomes a research focus in recent years. Many methods were proposed to tackle this problem, such as model pruning \cite{han2015deep,li2016pruning}, quantization \cite{Hubara2016Quantized, Wu2016Quantized} and knowledge transfer \cite{hinton2014distilling, romero2014fitnets}.

 \begin{figure}[htbp]
    \begin{center}
    \begin{minipage}{0.7\linewidth}
      \subfigure[Knowledge Distillation]{
        \hspace{8pt}
        \includegraphics[width=0.4\textwidth]{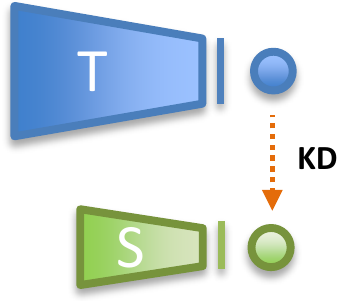}
        \hspace{8pt}
      }
      \subfigure[Hint Learning]{
        \includegraphics[width=0.4\textwidth]{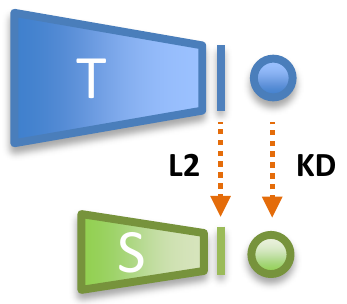}
      }
      \vspace{-10pt}
      \\ \vspace{-10pt}
      \subfigure[Route Constrained Hint Learning]{
        \includegraphics[width=0.99\textwidth]{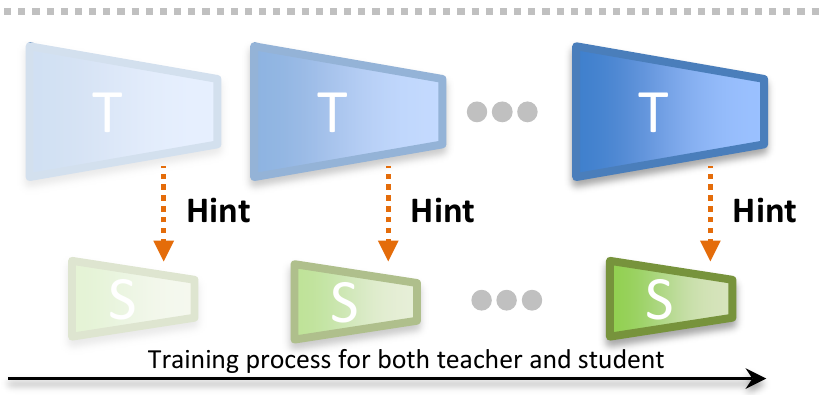}
      }
    \end{minipage}
    \begin{minipage}{0.25\linewidth}
      \includegraphics[width=0.95\textwidth]{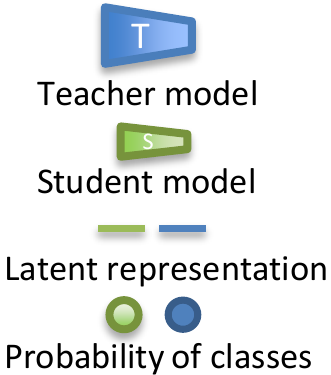}
    \end{minipage}
    \end{center}
      \caption{Comparing to targeting only a converged teacher like KD and Hint-based learning, RCO narrows the performance gap by gradually mimicking the route sequence of teacher.}
      \label{fig:schematic}
   \end{figure}

Among these approaches, knowledge distillation (KD) performs as an essential way to optimize a static model by mimicking the behavior (final predictions \cite{hinton2014distilling} or activations of hidden layers \cite{romero2014fitnets}) of a powerful teacher network as shown in Figure \ref{fig:schematic}(a) and Figure \ref{fig:schematic}(b). Guided by this softened knowledge, a student network could pay more attention to extra supervision such as the probability correlation between classes rather than the one-hot label.

Previous methods only consider single converged model as teacher to teach small student network, which may result that the student stucks in approximating teacher's performance along with the increasing gap (in capacity) between teacher and student \cite{mirzadeh2019improved}.
We observe that student supervised by teacher's early training stage has a much smaller performance gap with its teacher than that supervised by teacher's latter training stage. 
From the perspective of curriculum learning \cite{Bengio2009Curriculum}, a reasonable explanation beyond this observation is that teacher's knowledge is gradually becoming harder along with the training process. 
We claim that the intermediate states that teacher passed by are also valuable knowledge for easing the learning process and lowering the error bound of the student.

Based on this philosophy, we propose a new method called RCO which supervises student with the teacher's optimization route. Figure \ref{fig:schematic}(c) shows the whole framework of RCO. Comparing to single converged model, the route of teacher contains extra knowledge through providing an easy-to-hard learning sequence. By gradually mimicking such sequence, the student can learn more consistent with the teacher, therefore narrowing the performance gap. Besides, we analyze the impact of different learning sequence on performance and propose an efficient method based on greedy strategy for generating sequence, which can be used to shorten the training paradigm meanwhile maintaining high performance.

Extensive experiments on CIFAR-100, ImageNet-1K and large scale face recognition show that RCO significantly outperforms knowledge distillation and other SOTA methods on all the three tasks. Moreover, our method can be combined with previous knowledge transfer methods and boost their performance. To sum up, our contribution could be summarized into three parts:

\begin{itemize}
  \item
  We rethink the knowledge distillation model from the perspective of teacher's optimization path and learn the significant observation that learning from the converged teacher model is not the optimal way. 
  \item 
  Based on the observation, we propose a novel method named RCO that utilizes the route in parameter space that teacher network passed by as a constraint to lead a better optimization on student network.
  \item
  We demonstrate that the proposed RCO can be easily generalized to both knowledge distillation and hint learning. Under the same data and computational cost, RCO outperforms KD by a large margin on CIFAR, ImageNet and a one-to-million face recognition benchmark Megaface.
\end{itemize}

%------------------------------------------------------------------------

\section{Related Work}
\textbf{Neural Network Miniaturization.} 
Many works study the problem of neural network miniaturization. They could be categorized into two methods: designing small network structure and improving the performance of small network via knowledge transfer. As for the former, many modifications on convolution were proposed since the original convolution took up too many computation resources. MobileNet~\cite{howard2017mobilenets} used depth-wise separable convolution to build block, ShuffleNet~\cite{Zhang2017ShuffleNet} used pointwise group convolution and channel shuffle. These methods could maintain a decent performance without adding too much computing burden at inference time.
Besides, many studies \cite{han2015learning, Molchanov2016Pruning, liu2017learning, He2018AMC} focus on network pruning, which boosts the speed of inference through removing redundancy in a large CNN model. Han \etal~\cite{han2015learning} proposed to prune nonsignificant connections. Molchanov \etal~\cite{Molchanov2016Pruning} presented that they could prune filters with low importance, which were ranked by the impact on the loss. They approximated the change in the loss function with Taylor expansion. These methods typically need to tune the compression ratio of each layer manually. Most recently, Liu \etal~\cite{liu2017learning} presented the network slimming framework. They constrained the scale parameters of each batch normalization \cite{ioffe2015batch} layer with sparsity penalty such that they could remove corresponding channels with lower scale parameters. In this way, unimportant channels can be automatically identified during training and then removed. He \etal~\cite{He2018AMC} proposed to adopt reinforcement learning to exploit the design space of model compression. They benefited more from replacing manually tuning with automatical strategies.

As for the latter, the most two popular knowledge transfer methods are Knowledge Distillation~\cite{hinton2014distilling} and FitNet~\cite{romero2014fitnets}. We mainly consider these situations in this work.

\textbf{Knowledge Distillation for Classification.} % 
Efficiently transferring knowledge from large teacher network to small student network is a traditional topic which has drawn more and more attention in recent years.
Caruana \etal~\cite{Caruana2006Model} advocated it for the first time. They claimed that knowledge of an ensemble of models could be transferred to the other single model. Then Hinton \etal~\cite{hinton2014distilling} further claimed that knowledge distillation (KD) could transfer distilled knowledge to student network efficiently. By increasing the temperature, the logits (the inputs to the final softmax) contain richer information than one-hot labels.
KD \cite{hinton2014distilling} is generally used for close-set classification, where the training set and testing set have exactly the same classes.

\textbf{Learning Representation from Hint.} 
Hint-based learning is often used for open-set classification such as face recognition and person Re-identification.
FitNet~\cite{romero2014fitnets} firstly introduced more supervision by exploiting intermediate-level feature maps from the hidden layers of teacher to guide training process of student. Afterward, Zagoruyko \etal~\cite{Zagoruyko2016Paying} proposed the method to transfer attention maps from teacher to student. Yim \etal~\cite{yim2017gift} defined the distilled knowledge from teacher network as the flow of the solution process (FSP), which is calculated by the inner product between feature maps from two selected layers.

Previous knowledge transfer methods only supervise student with converged teacher, thus fail to capture the knowledge during teacher's training process. Our work differs from existing approaches in that we supervise student with the knowledge transferred from teacher's training trajectory.

\section{Route Constrained Optimization}
\begin{figure*}[htbp]
    \begin{center}
      \includegraphics[width=0.9\textwidth]{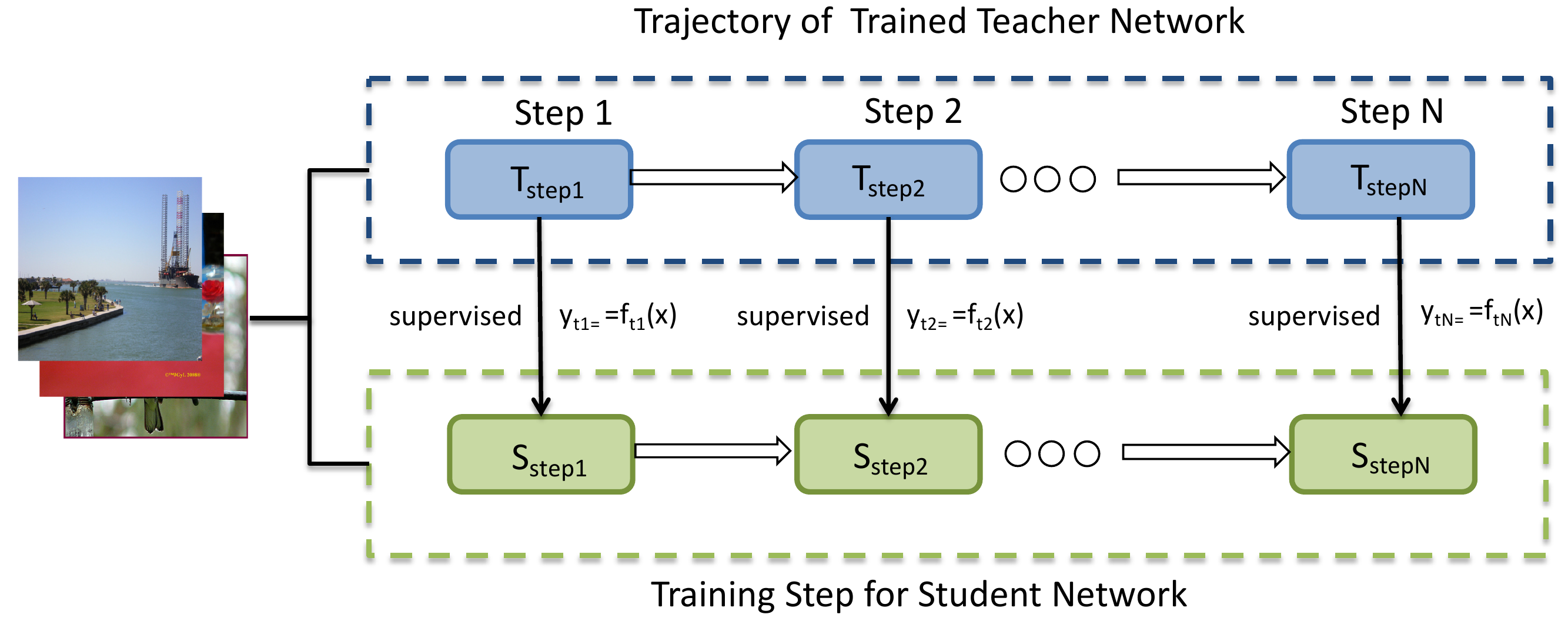}
    \end{center}
      \caption{The overall framework of RCO. Previous knowledge transfer method only considers the converged teacher model. While RCO aims to supervise student with intermediate training state of teacher. 
       }
      \label{fig:pipeline}
   \end{figure*}

\subsection{Teacher-Student Learning Mechanism}
For better illustration, we refer teacher network as $\phi_t$ with parameters $W_t$, and student network as $\phi_s$ with parameters $W_s$. $\ P_t=softmax(z_t)$ and $\ P_s=softmax(z_s)$ represent output predictions of teacher and student respectly, $z_t$ and $z_s$ for logits of teacher and student.
The idea beyond KD is to let the student mimic the behavior of teacher by minimizing the cross-entropy loss and Kullback–Leibler divergence between predictions of teacher and student as follows:
\begin{equation}
L_{KD} = H(y, P_s) + \lambda KL(P_t^\tau, P_s^\tau),
\end{equation}
\noindent where $\tau$ is a relaxation hyperparameter (referred as Temperature in \cite{hinton2015distilling}) for softening the output of teacher network, and $\lambda$ is a hyper-parameter for balancing cross-entropy and KL divergence loss.
In several works \cite{Gregor2016Do,li2017mimicking} the KL divergence is replaced by euclidean distance, 
\begin{equation}
L_{\text{mimic}} = \frac{1}{n} \sum_{i=1}^n  \left \|\bm{f}_s - \bm{f}_t\right\|_2^2, \label{mimic_loss}
\end{equation}
\noindent $f$ represents the feature representations.

\subsection{Difficulty in Optimizing Student} \label{sec: difficulty of current method}
In common, the teacher invokes a larger and deeper network for arriving at a lower local minimum and achieving higher performance. It is hard for a smaller and shallower student to mimic such a large teacher due to the huge gap (in capacity) between teacher and student. 
Usually, the network is trained to minimize the objective function by using stochastic gradient descent. Due to the high non-convex of the loss function, there are many local minima in the training process of deep neural networks. When the network converges to a certain local minimum, its training loss will converge to a certain (or similar) value regardless of different initializations.

\begin{table}[htbp]
\begin{center} % used for centering table
\scalebox{0.9}{
\begin{tabular}{c | c |c | c | c | c } % centered columns (4 columns)
\toprule[1pt]      %inserts double horizontal lines
Network & Epoch & 10 & 40 & 120 & 240 \\ 
\hline
\multirow{2}{*}{ResNet-50} & T top-1(\%) & 53.07 & 56.53 & 77 & 79.52 \\
  & T loss & 1.680 & 1.199 & 0.067 & 0.009 \\
\hline
\multirow{2}{*}{MobileNetV2} & S top-1(\%) & 51.21 & 57.62 & 66.05 & 68.71 \\      
  & S loss & 0.511 & 1.189 & 3.758 & 4.218 \\
  
\bottomrule[1pt] 
\end{tabular}
}

\end{center}
%      \captionsetup[table]{skip=5pt}
\caption{The performance of student network trained with different epochs from teacher's training trajectory on CIFAR-100 dataset. ``T'' and ``S'' stand for teacher and student separately. ``loss'' represents training loss.} 
\label{table:mimic_objective} 
\end{table}

\textbf{Could we reach a better local-minimal than this condition?} 
We consider changing the optimization objective.
% As the teacher becomes more and more deterministic during the training process,
More specifically, the student is trained by mimicking less deterministic target first, then moving forward to deterministic one, Hoping in this way the student has smaller gap with teacher. 
To validate this, we use different intermediate states of teacher to supervise student, and use the training loss and top-1 accuracy to evaluate the difference between target teacher and converged student. MobileNetV2 \cite{sandler2018mobilenetv2} is adopted as student and ResNet-50 \cite{He2016Deep} as teacher. The teacher network is trained by cross-entropy loss, the student network is trained by KD loss. We select checkpoints of teacher at 10th epoch, 40th epoch, 120th epoch, and 240th epoch as the target to train student network separately. The checkpoint at 240th epoch is the final converged model, and checkpoint at 10th epoch is least deterministic in the analysis.

Table \ref{table:mimic_objective} summarizes the results. 
It can be observed from the table that the student guided by the less deterministic target has lower training loss, and more convergent target brings larger gap in performance.
In other words, the more convergent teacher means a harder target for student to approach.

Inspired by curriculum learning \cite{Bengio2009Curriculum} that the local minima can be promoted by the easy-to-hard learning process, we take the sequence of teacher's intermediate states as the curriculum to help the student reach a better local minimum. 

\subsection{RCO}
For better illustration, we refer the intermediate training states (checkpoints) used to form the learning sequence as \textbf{anchor points}. Suppose there are $n$ anchor points on the teacher's trajectory. The overall framework of RCO is shown in Figure \ref{fig:pipeline}. 

Without loss of generality, let $\bm{C} = C_1, C_2, ..., C_n$ represent the anchor points set, and the correpsonding outputs are $ \phi_t(x; W_{C_1}), \phi_t(x; W_{C_2}), ..., \phi_t(x; W_{C_n}) $ . The training for student is started from random initialization. Then we train the student step-by-step to mimic the anchor point on teacher's trajectory until finishing training with the last anchor point.
At $i_{th}$ step, the learning target of student is switched to the output $\phi_t(x; W_{C_i})$ of $i_{th}$ anchor point. The optimization goal of $i_{th}$ step is as follows:
\begin{equation}
Loss_i = H(y, \phi_s(x; W_s)) + \lambda H(\phi_s(x; W_s), \phi_t(x; W_{C_i})), 
\end{equation}
where $i \in \{1,2,...,n\}$. The parameter \bm{$W_s$} is optimized by learning to these anchor points sequentially. Algorithm \ref{algo:first} describes the details of the whole training paradigm.

\begin{algorithm}
\caption{Route Constrained Optimization}
\label{algo:first}
\begin{algorithmic} 
\REQUIRE anchor points set from pre-trained teacher network: $C_1, C_2,..., C_n$, student network with parameter $W_i$
\STATE $i=1$
\STATE Randomly initialize $W_i$
\WHILE{$i \leq n$}
    \STATE Initialize teacher network with $C_i$ anchor, get $W_{C_i}$
  \IF{$i > 1$}
      \STATE Initialize $W_i$ with $W_{i-1}$
  \ENDIF
  % \REPEAT 
  \STATE update the $W_i$ by optimizing $L_{KD}(W_i, W_{C_i})$
  % \UNTIL Convergence, get $W_S^i$
  \STATE $i=i+1$
\ENDWHILE
\STATE get $W_n$ as the final weights of student.
\end{algorithmic}
\end{algorithm}

\subsection{Rationale for RCO}
From the perspective of curriculum learning, the easy-to-hard learning sequence can help the model get a better local minimum \cite{Bengio2009Curriculum}. RCO is similar to curriculum learning but different in that it provides an easy-to-hard labels sequence on teacher's trajectory.

Let $\mathcal{Y}_i$ be the output of $i_{th}$ anchor point. The outputs of whole anchor points constrcut the space $ \Omega=\{\mathcal{Y}_i|i=1,2,..., n\}$. The results shown in table \ref{table:mimic_objective} premises that the intermediate states on teacher's trajectory construct an easy-to-hard sequence, eg. $\mathcal{Y}_i$ is easier to mimic than $\mathcal{Y}_{i+1}$ while $\mathcal{Y}_n$, the converged model, is the hardest objective for a small student.

Let the $\mathcal{X}$ be the training data. The training data and output of $i_{th}$ anchor pair $(\mathcal{X}, \mathcal{Y}_i)$ provide a lesson. Then the curriculum sequence can be formulated as follows:
\begin{equation}
  \{(\mathcal{X}, \mathcal{Y}_{i})|i = 1, ..., n\}.
\end{equation}

Without loss of generality, let $\mathcal{L}_\lambda(\mathcal{X};\theta)$ represents a single-parameter family of cost functions such that $L_1$ can be easily optimized, while $L_N$ is the criterion that we actually wish. During the sequential training of RCO, increasing $\lambda$ means adding the hardness of learning through switching anchor points. 
Let $\mathcal{D}$ represent the hardness metric for a learning target. As shown in Section \ref{sec: difficulty of current method}, more convergence of anchor means more hardness of learning target,  
\begin{equation}
\mathcal{D}( \phi( \mathcal{X}, W_{C_i}) < \mathcal{D}( \phi( \mathcal{X}, W_{C_{i+1}})~~~\forall{i} > 0.
\end{equation}
In curriculum learning \cite{Bengio2009Curriculum} the sequence of learning is generated by spliting the $\mathcal{X}$ in to several ``lessons'' with different hardness depending on a predefined criterion. While RCO can be seen as a more flexible approach, which gradually changes the hardness of target labels $\mathcal{Y}$. Both curriculum learning and RCO work by easy-to-hard learning to move $\theta$ gradually into the basin of attraction of a dominant (if not global) minimum \cite{Bengio2009Curriculum}.

\subsection{Strategy for Selecting Anchor Points}  \label{sec:strategy}

\textbf{Equal Epoch Interval Strategy.} Typically, the teacher network could produce tremendous checkpoints during the training process. To find the optimal learning sequence, one can search it with brute force. However, given $n$ possible intermediate states, there are $2^n$ possible sequences, which is impractical to implement.
A straightforward strategy is supervising the student by every state (epoch/iteration) on teacher's trajectory. However, mimicking every state is dispensable and time-consuming since adjacent training states are very close to each other. Given limited time, a more efficient way is to sample epochs with equal epoch interval (\textbf{EEI}), eg. select one for every four epochs. 

Although efficient in time, EEI is a quite simple ad-hoc method that ignores the hardness between different anchor points, and it would lead to an improper curriculum sequence. The desirable property of the curriculum sequence should be efficient to quickly learn and smooth in hardness to better bridge the gap between teacher and student.

\begin{figure}
    \begin{center}
    \includegraphics[width=0.48\textwidth, angle=0]{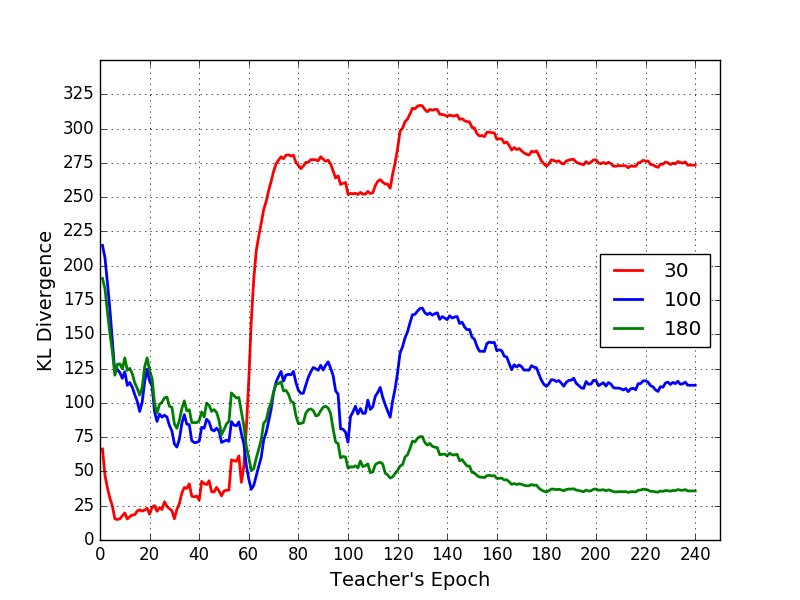}
  \end{center}
    \caption{The curve of KL divergence loss between student supervised by teacher's different epochs (30th, 100th, 180th) and teacher's all 240 epochs on validation set.}
    \label{fig:epoch-10}
\end{figure}

\textbf{Greedy Search Strategy.} To delve into optimization route of student when learning to teacher, we count the KL divergence between outputs of student and different target states of teacher on validation set, which consists of 10k examples random sampled from training set. The teacher is trained by dropping learning rate at 60th,120th,180th epochs separately. We choose 30th, 100th, and 180th epochs as target states to supervise the student separately. Figure \ref{fig:epoch-10} shows the KL divergences curve between student and intermediate states of teacher. From the figure we can observe that the student supervised by teacher's 30th epoch is very close with teacher's 30th epoch, but has large gap with teacher's latter epochs, especially after teacher dropping learning rate. The same observations can be found from student supervised by teacher's other epochs.

Table \ref{table:mimic_objective} and Figure \ref{fig:epoch-10} give us two insights: a particular student has the ability to learn harder target limited in a certain range; supervised by a better teacher would promote the ability of student. 

Inspired by these insights, we propose a greedy search strategy (\textbf{GS}) to find efficient and hardness-smooth curriculum sequence. The goal of greedy strategy is to find the one which is on the boundary of range that student can learn. To find those boundary anchor points, a metric is introduced as follows:
\begin{small}
\begin{equation} \label{ratio_define}
    \begin{split}
      r_{ij} &= \frac{\mathcal{H}_j - \mathcal{H}_i}{\mathcal{H}_i} , i,j\in\{i+1,i+2,...,N\}, \\ 
      \mathcal{H}_i &= \mathcal{H}( \phi_s(\mathcal{X^{'}}, W_s), \phi_t(\mathcal{X^{'}}, W_{t_i}), \\
      \mathcal{H}_j&=\mathcal{H}( \phi_s(\mathcal{X^{'}}, W_s), \phi_t(\mathcal{X^{'}}, W_{t_j}), \\
    \end{split}
  \end{equation}
\end{small}
\noindent where $\mathcal{H}$ is the KL divergence, and $\mathcal{X^{'}}$ is the validation set. $r_{ij}$ evaluates the hardness of $j_{th}$ epochs of teacher for a student guided by $i_{th}$ epochs. Then we refer a hyper-parameter $\delta$ as the threshold which indicates the learning ability of student. When $r_{ij}>\delta$, it means that $j_{th}$ epoch is hard for student trained by $i_{th}$ to learn, and $r_{ij}<\delta$ means inverse. 
Based on above philosophy, we give the complete \textbf{GS} strategy in Algorithm \ref{algo:second}.

It seems the anchor points that near the point of tuning learning rate are more important than other anchor points. Intuitively,  according to Algorithm \ref{algo:second}, the optimal learning sequence must contain at least one anchor point from different learning rate stage. 
Since this section mainly focuses on the strategy for anchor points selection, we provide the empirical value of $\delta=0.8$ for MobileNetV2 to achieve a better balance between performance and training cost.
Note that although our experiments are based on SGD, GS is also applicable for other optimization methods like SGDR\cite{loshchilov1608sgdr}, since prerequisites are still true.

\begin{algorithm}
\caption{Greedy Search}
\label{algo:second}
\begin{algorithmic} 
\REQUIRE Student network with parameter $W_s$ after mimicking former $i_{th}$ anchor point $C_i$, where $i\in \{1,2,...,N\}$, relaxation factor $\delta$.
  \STATE compute KL divergence $\mathcal{H}_i$ 
  \STATE $j=i+1$
  \WHILE{$j < N$}
        \STATE compute $\mathcal{H}_j$ on validation set
        % \STATE compute $r_{ji}=(D_{kj}-D_{ki})/D_{ki}$
        \STATE compute $r_{ij}=\frac{\mathcal{H}_j - \mathcal{H}_i}{\mathcal{H}_i}$
        \IF{$r_{ij} > \delta$}
        %  \STATE Initialize $W_S$ with $W_S^{i-1}$
          \STATE Return j-1;
        %\Return your-text

        \ENDIF
            \STATE $j=j+1$
  \ENDWHILE
  \STATE Return N;
\end{algorithmic}
\end{algorithm}

\section{Experiments} \label{sec:experiments}
\textbf{Common Settings.}
The backbone network for teacher in all experiments is ResNet-50. For the student structure, instead of using smaller ResNet, we use more compact MobileNetV2 as well as its variants with different FLOPs, since MobileNetV2 has proven to be highly effective in keeping high accuracy while maintaining low FLOPs in many tasks.
Expansion ratio and width multiplier are two tunable parameters to control the complexity of the MobileNetV2. We make default configuration by setting expansion ratio to 6 and width multiplier to 0.5. The relaxation is 5 in KD loss. Note that all these experiments are based on GS that usually produces about 4 anchor points.

\subsection{Experiment on CIFAR-100}  \label{sec:exp-cifar}

% \subsubsection{Experiment}
The CIFAR-100 dataset contains 50 000 images in training set and 10 000 images in validation set with size 32$\times$32. In this experiment, for the teacher network we set initial learning rate to 0.05 and divide it by 10 at 150th, 180th, 210th epochs and we train for 240 epochs. We set weight decay to 5e-4, batch size to 64 and use SGD with momentum. For the student network, the setting is almost identical with teacher's except that the initial learning rate is 0.01.

We compare the top-1 accuracy of CIFAR-100 dataset and show the result in Table \ref{table:CIFAR-100}. From the result we can find that our method improves about 2.1\% on top-1 compared with KD.

\begin{figure}
\begin{center}
 \includegraphics[width=0.48\textwidth, angle=0]{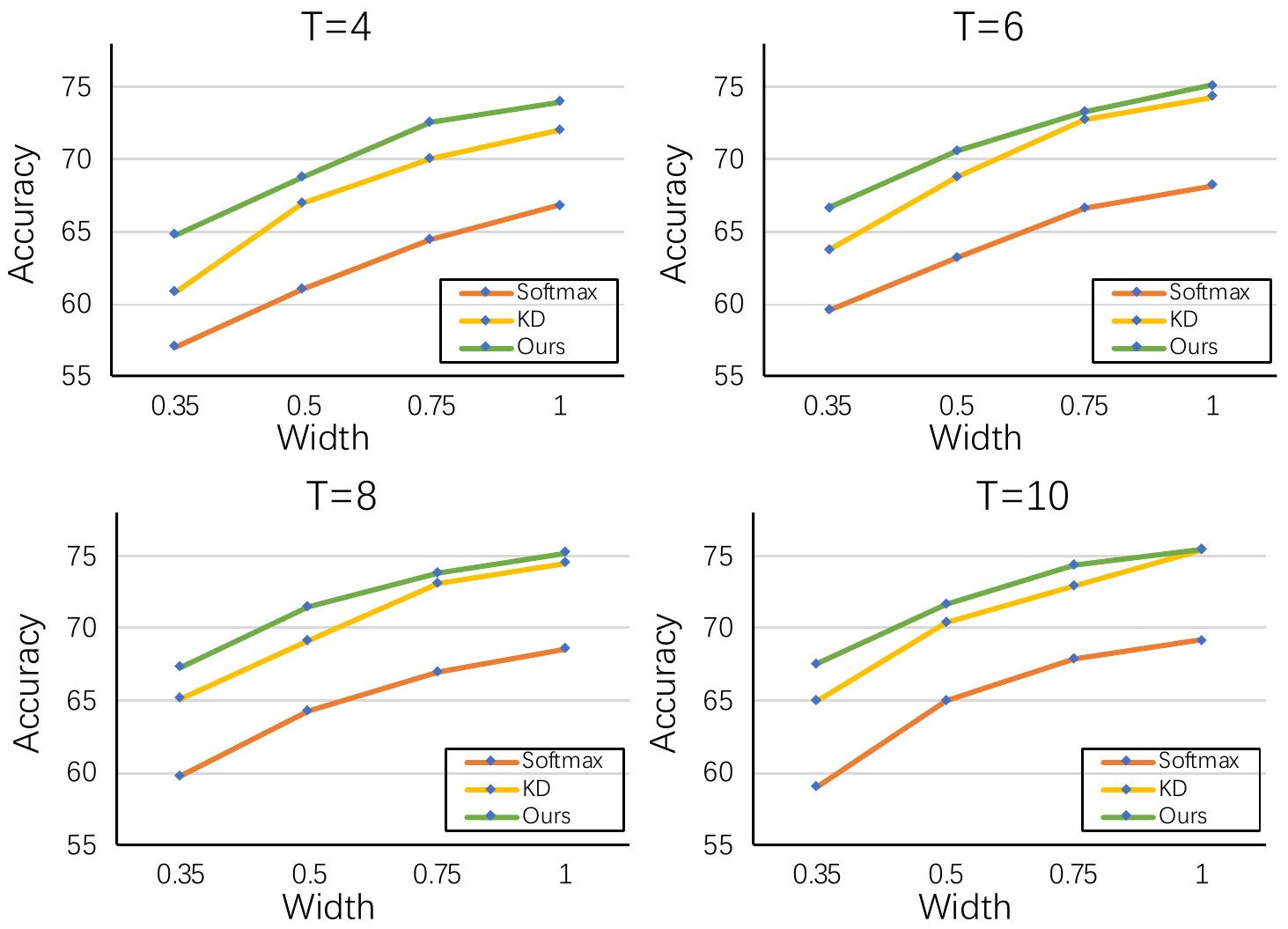}
  \end{center}
 \caption{CIFAR-100 top-1 accuracy of MobileNetV2 with different settings. The ``Width'' in $x$-axis represents width multiplier and ``T'' in the title stands for the expansion ratio. The proposed method acquires more promotion with smaller student network.}
 \label{fig:flops}
\end{figure}

%   expansion ratio and width multipliers
Although the base student network is small and fast, it is common that some specific cases or applications require the model to be smaller and faster. To further investigate the effectiveness of the proposed method, we conduct extensive experiments by applying RCO to a series of MobileNetV2 with different width multipliers and expansion ratios. We set expansion ratio to 4, 6, 8, 10 and width multiplier to 0.35, 0.5, 0.75, 1.0 separately, which forms totally 16 different combinations. The FLOPs of these models are shown at Table \ref{table:flops-table}. We rank the model according to the width multiplier and draw the result on Figure \ref{fig:flops}. From the figure we can make the following observations: (i) The proposed method exhibits consistently superiority in all settings. (ii) The student network with smaller capacity(e.g. MobileNetV2 with T=4, Width=0.35) generally gains more improvement from RCO. (iii) Although the model with expansion ratio set to 10 and width multiplier set to 0.35 has larger FLOPs than the model with expansion ratio set to 4 and width multiplier set to 0.5, the former setting shows performance reduction among all three methods. It indicates that parameterizing expansion ratio with 10 and width multiplier to 0.35 largely limits representation power.
\begin{table}[ht]
    \begin{center} % used for centering table
    \begin{tabular}{c | c | c | c | c} % centered columns (4 columns)
    \toprule[1pt] %inserts double horizontal lines
    Method & Network & MFlops & top-1 & Loss  \\ 
    \hline
    T-Softmax & ResNet-50 & 2.6k & 79.34 & - \\\hline
    S-Softmax & MobileNetV2 & 13.5 & 61.88 & - \\
    S-KD & MobileNetV2 & 13.5 & 68.71 & 1.59 \\
    S-RCO & MobileNetV2 & 13.5 & \textbf{70.85} & 1.45 \\
    \bottomrule[1pt]
    \end{tabular}
  \end{center}
    \caption{Results on CIFAR-100} % title of Table
    \label{table:CIFAR-100} 
    \end{table}

\begin{table}[ht]
    \begin{center} % used for centering table
%      \begin{tabular}{| c | c | c | c | c |} % centered columns (4 columns)
    \begin{tabular}{ c |  c  c  c  c } % centered columns (4 columns)
    \toprule[1pt] %inserts double horizontal lines
    \multirow{2}{*}{Expansion ratio} & \multicolumn{4}{|c}{Width multiplier} \\ \cline{2-5} 
     & 0.35 & 0.5 & 0.75 & 1.0  \\ \hline
    4 & 5.4 & 9.8 & 19.3 & 32.1 \\
    6 & 7.3 & 13.5 & 27.2 & 45.6 \\
    8 & 9.1 & 17.1 & 35 & 59.1 \\
    10 & 11 & 20.8 & 42.8 & 72.6 \\ \bottomrule[1pt]
               
    \end{tabular}
  \end{center}
    \caption{Complexity (MFLOPs) for MobileNetV2 with different settings} % title of Table
    \label{table:flops-table} 
\end{table}

\subsection{Experiment on ImageNet} \label{sec:exp-imagenet}
The ImageNet dataset contains 1000 classes of images with various sizes. It is the most popular dataset in classification task. In this experiment, for the training of teacher network, we set initial learning rate to 0.4 and drop by 0.1 at 15k, 30k and 45k iterations and we train for 50k iterations. We set weight decay to 5e-4, batch size to 3072 and use SGD with momentum. 
As for the student network, we set initial learning rate to 0.1 and drop by 0.1 at 45k, 75k and 100k iterations and we train for 130k iterations. We set weight decay to 5e-4, batch size to 3072 and use SGD with momentum. In order to keep training stable, we use warm-up suggested by \cite{goyal2017accurate}  when training with large batch size.
We compare the top-1 and top-5 accuracy of ImageNet dataset and show the result in Table \ref{table:ImageNet}. From the result we can find that our method improves about 1.5\%/0.7\% on top-1/top-5 compared with KD, which verifies that RCO is applicable to large-scale classification.
\begin{table}[ht]
\begin{center} % used for centering table
\begin{tabular}{c | c | c c} % centered columns (4 columns)
\toprule[1pt] %inserts double horizontal lines
Method & Network & top-1 & top-5 \\ 
\hline
Teacher-Softmax & ResNet-50 &75.49 & 92.48 \\\hline
Student-Softmax & MobileNetV2 &64.2 & 85.4 \\
Student-KD & MobileNetV2 &66.75 & 87.3 \\
Student-RCO & MobileNetV2 & \textbf{68.21} & \textbf{88.04} \\
\bottomrule[1pt]
\end{tabular}
  \end{center}
%\captionsetup[table]{skip=5pt}
\caption{Results on ImageNet} % title of Table
\label{table:ImageNet} 
\end{table}

\subsection{Experiment on Face Recognition} \label{sec:exp-megaface}

Unlike classification, the network in face recognition usually contains a feature layer implemented as a fully-connected layer to represent the projection of each identity. Empirical evidence \cite{Luo2016Face} shows that mimicking the feature layer as the way used in FitNet \cite{romero2014fitnets} could bring more improvements for student network. We follow this setting in our baseline experiments.

We take two popular face recognition datasets MS-Celeb-1M \cite{guo2016ms} and IMDb-Face \cite{wang2018devil} as our training set and validate our method on MegaFace.
The \textbf{MS-Celeb-1M}  is a large public face dataset which contains one million identities with different age, sex, skin, and nationality and is widely used in face recognition area.
The \textbf{IMDb-Face} dataset contains about 1.7 million faces, 59k identities. All images are obtained from IMDb website. The \textbf{MegaFace} is one of the most popular benchmarks that could perform face recognition under up to 1 million distractors. This benchmark is evaluated through probe and gallery images from FaceScrub.

In this experiment, for the teacher network, we set initial learning rate to 0.1 and drop by 0.1 at 100k, 140k, 170k iterations and we train for 220k iterations. We set weight decay to 5e-4, batch size to 1024 and use SGD with momentum. We resize the input image to 224$\times$224 without augmentation. We use ArcFace \cite{Deng2018ArcFace} to train the teacher network.
As for the student network, we set initial learning rate to 0.05 and drop by 0.1 at 180k, 210k iterations and we train for 240k iterations. The rest settings are identical to the teacher.

We show our result on Table \ref{table:MegaFace}. From the table we can see that on this challenging face recognition task, RCO largely boosts the performance of MobileNetV2 \cite{sandler2018mobilenetv2} compared to original hint-based learning.
\begin{table}[ht]
\setlength{\tabcolsep}{5pt}
\begin{center} % used for centering table
\begin{tabular}{c | c c c c c c} % centered columns (4 columns)
\toprule[1pt] %inserts double horizontal lines
    & \multicolumn{6}{|c}{top-1 @ distractor size} \\
\hline
Method & $e^1$  & $e^2$ & $e^3$ & $e^4$ & $e^5$ & $e^6$ \\ 
\hline 
Teacher & 99.78 & 99.67 & 99.38 & 98.86 & 97.70 &  94.83 \\\hline
Softmax & 99.20 & 96.37 & 91.49 & 84.45 & 75.60 &  65.91 \\ 
FitNet & 99.62 & 98.80 & 96.83 & 93.53 & 88.28 &  81.02 \\ 
RCO & \textbf{99.69} & \textbf{99.01} & \textbf{97.52} & \textbf{94.84} & \textbf{90.55} &  \bf{84.3} \\
\bottomrule[1pt]
\end{tabular}
  \end{center}
%\captionsetup[table]{skip=5pt}
\caption{Results on MegaFace} % title of Table
\label{table:MegaFace} 
\end{table}

\subsection{Ablation Studies}   \label{sec:sensitivity of anchor points}

Although RCO achieves decent result in previous experiments, the extra training time that it brings is not negligible. Even if we just construct the learning sequence with 4 anchor points, it still needs 4 times training epochs compared with KD or Softmax.
Since training time plays an important role in either research or industrial, 
we consider using the same time as KD to verify the robustness of RCO.
Note that we set expansion ratio to 4 and width multiplier to 0.35 for the backbone MobileNetV2 in this section.

% \subsubsection{Comparison with Other SOTA Methods}

\textbf{Comparison under Limited Training Epochs.} Previous experiments commonly need more training epochs than KD. Consider performing RCO with EEI strategy on CIFAR-100. Let \textbf{$M_{gap}$} be the epoch interval used in EEI. To get 4 anchor points, we can set $M_{gap}$ to 60. Then the selected anchor points should be 60th, 120th, 180th, and 240th epoch.
The student trained with the learning sequence needs 960 epochs in total since each anchor point is trained for 240 epochs to ensure convergence.

We then speed up the EEI strategy from multi-stage to one stage (\textbf{one-stage EEI}), where we only train student for 240 epochs, by simply modifying the training paradigm as follows:
the student is initially supervised by 60th epoch of teacher for the student's first 60 epochs, then supervised by teacher's 120 epoch for the next 60 epochs, and so on. 
% Training paradigm of EEI is modified as follows: 

In one-stage EEI, it is natural to evaluate the impact of different number of anchor points. 
Let $K$  be the size of training set. We start the $M_{gap}$ from the smallest case, where $M_{gap}$ is 1/($K$ /$BatchSize$) (It is 1.28E-3 in Table \ref{table:num_anchor}, which means student mimic teacher's every iteration). Then gradually increase $M_{gap}$ to the largest case, where $M_{gap}$ is the maximum epoch (240) and RCO degrades into KD.

From the perspective of optimization route, we find method in \cite{zhou2017rocket} could be regarded as a particular case of RCO when setting $M_{gap}$ to the smallest value, and matching logits with KD loss instead of MSE loss. Besides, We also follow \cite{zhang2017deep} to implement DML and make a comparison with KD.
% We present the result at Table \ref{table:num_anchor}. 
The result in Table \ref{table:num_anchor} shows that RCO outperforms other methods in all settings. By properly selecting $M_{gap}$ to 10, RCO gets 4.2\% and 3.8\% improvement on CIFAR-100 compared with KD and DML respectively.

\begin{table}[ht]
    \begin{center} % used for centering table
    \scalebox{0.9}{
    
    \begin{tabular}{c | c | c | c} % centered columns (4 columns)
    \toprule[1pt]      %inserts double horizontal lines
    %Epoch Interval & $M$ & 1 & 2 & 4 & 10 & 20 & 240 & 10(random)\\ 
    Method& $M_{gap}$ & Anchor Number & top-1\\
    \hline
  %   \multirow{8}{*}{Number of anchor}   &
    % 1.28E-3(DML) & 187500 & 61.13 \\
   DML\cite{zhang2017deep} & - & - & 61.13 \\
   RL\cite{zhou2017rocket} & 1.28E-3 & 187500 & 61.63 \\\hline
   \multirow{7}{*}{\textbf{RCO}} 
    & 1 & 240 & 62.74 \\
    & 2 &  120 & 63.78 \\
    & 4 & 60 & 64.21 \\
    & \textbf{10} & \textbf{24} & \textbf{65.01} \\
    & 20 & 12 & 63.88 \\
    & 60 & 4 & 64.5 \\\hline
    KD & 240 & 1 & 60.79 \\

    \bottomrule[1pt]
    \end{tabular}
    }
  \end{center}
    %\captionsetup[table]{skip=5pt}
    \caption{Comparison of RCO based on One-stage EEI with other knowledge transfer methods under limited training epochs. It clearly shows RCO outperforms other methods by using same training epochs.
    } % title of Table
    \label{table:num_anchor} 
\end{table}

% \subsubsection{Comparison on Different Strategies} \label{sec:compare strategy}
\textbf{Comparison on Different Strategies.} 
Since the strategy is the most crucial part of RCO, we make a comparison between these strategies. For practical considerations, we limit the training epoch to no more than four times the epochs of KD.
We have chosen the following strategies to compare: \textbf{one-stage EEI, EEI-x, GS}, where the ``x'' in ``EEI-x'' represents the number of selected anchor points with EEI strategy.
%For the ``Random'' strategy, we just randomly select the 10 anchor points and train them sequentially in one stage just as the paradigm used in one-stage EEI. 
%For the ``fast GS strategy'', if we get $N$ learning targets from GS, then we can compress the entire training epochs for each target to $1/N$ of original epochs and keep the whole training time the same as KD.
The results are shown in Table \ref{table:strategies}. 
From the result we make the following observations: 1) all strategies show great superiority to KD, 2) GS is the best strategy among them, thus should be used when training time is not a constraint.

\begin{table}[ht]
  \begin{center} % used for centering table
  \scalebox{0.9}{
  
  \begin{tabular}{ c| c | c | c | c} % centered columns (4 columns)
  \toprule[1pt]      %inserts double horizontal lines
  Strategy & one-stage & $M_{gap}$ & Total Epoch & top-1\\
  \hline
  KD & \ding{51} & 240 & 240 & 60.79 \\
  \textbf{one-stage EEI} & \ding{51} & 10 & 240 & \textbf{65.01} \\\hline
  EEI-2 & \ding{55} & 120 & 480 & 61.43 \\
  EEI-3 & \ding{55} & 80 & 720 & 63.34 \\
  EEI-4 & \ding{55} & 60 & 960 & 65.27 \\
%   Random  &  240 & 61.9 \\
%   fast Greedy &  240 & xx.xx \\
  \textbf{GS} & \ding{55} & - & 720 & \textbf{65.41} \\
  \bottomrule[1pt]
  \end{tabular}
  }
\end{center}
  %\captionsetup[table]{skip=5pt}
  \caption{Comparison of RCO based on different strategies on CIFAR-100. GS achieves the best result among all strategies.
  } % title of Table
  \label{table:strategies} 
\end{table}

\subsection{Visualization}
\begin{figure}
    \begin{center}
    \includegraphics[width=0.4\textwidth, angle=0]{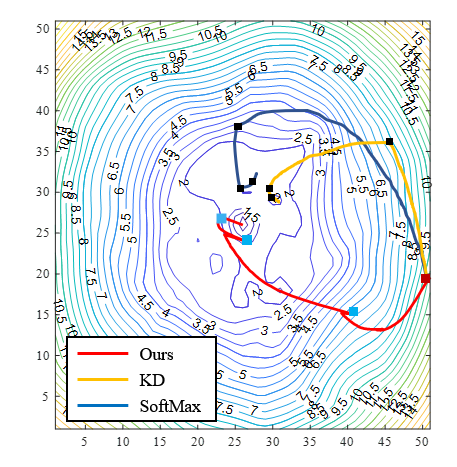}
  \end{center}
    \caption{Visualization of student's training trajectories by PCA directions for MobileNetV2 on CIFAR-100 dataset. These student networks are trained by different methods (SoftMax, KD, and Ours) and initialized with the same random parameters. 
      The red curve stands for student trained with RCO, the blue dot on the line represents the location guided by intermediate anchor point. The red curve arrives at the lowest local minimum among them.}
    \label{fig:loss_pca}
\end{figure}

\textbf{Visualization of Trajectory.} In order to further analyze our method, we plot student's training trajectory using PCA directions suggested by Li \etal~\cite{li2017visualizing}. The process is as follows:
Given $n$ training epochs, let $W_{m_i}$ denote model parameters at epoch $i$ and the final estimate as $W_{m_n}$.
Then we apply PCA to the matrix $[W_{m_0}-W_{m_n};...;W_{m_{n-1}}-W_{m_n}]$ and choose the most two principal directions. 

In Figure \ref{fig:loss_pca} the training trajectory of MobileNetV2 on CIFAR-100 is plotted for student in three modes: 1) Softmax, 2) KD, 3) the proposed method (Ours). 
For a fair comparison, three students are initialized with the same parameters (marked as red dot) and are trained for 240 epochs each, where RCO uses one-stage EEI with three anchor points. For the curve of RCO, the blue dots on the line show the epochs where student is guided by intermediate anchor points. For KD or Softmax, epochs where the learning rate was decreased are shown as black dots.

The first anchor point keeps the student away from the direction suggested by Softmax or KD and arrives at an intermediate state. The state itself may not lie in a well-performed parameter space, but with the guidance of succeeding anchor points the student network eventually reaches to a deeper local minimum, which has adequately demonstrated the importance of optimization route from teacher network.

\begin{figure}
 \includegraphics[width=0.45\textwidth, angle=0]{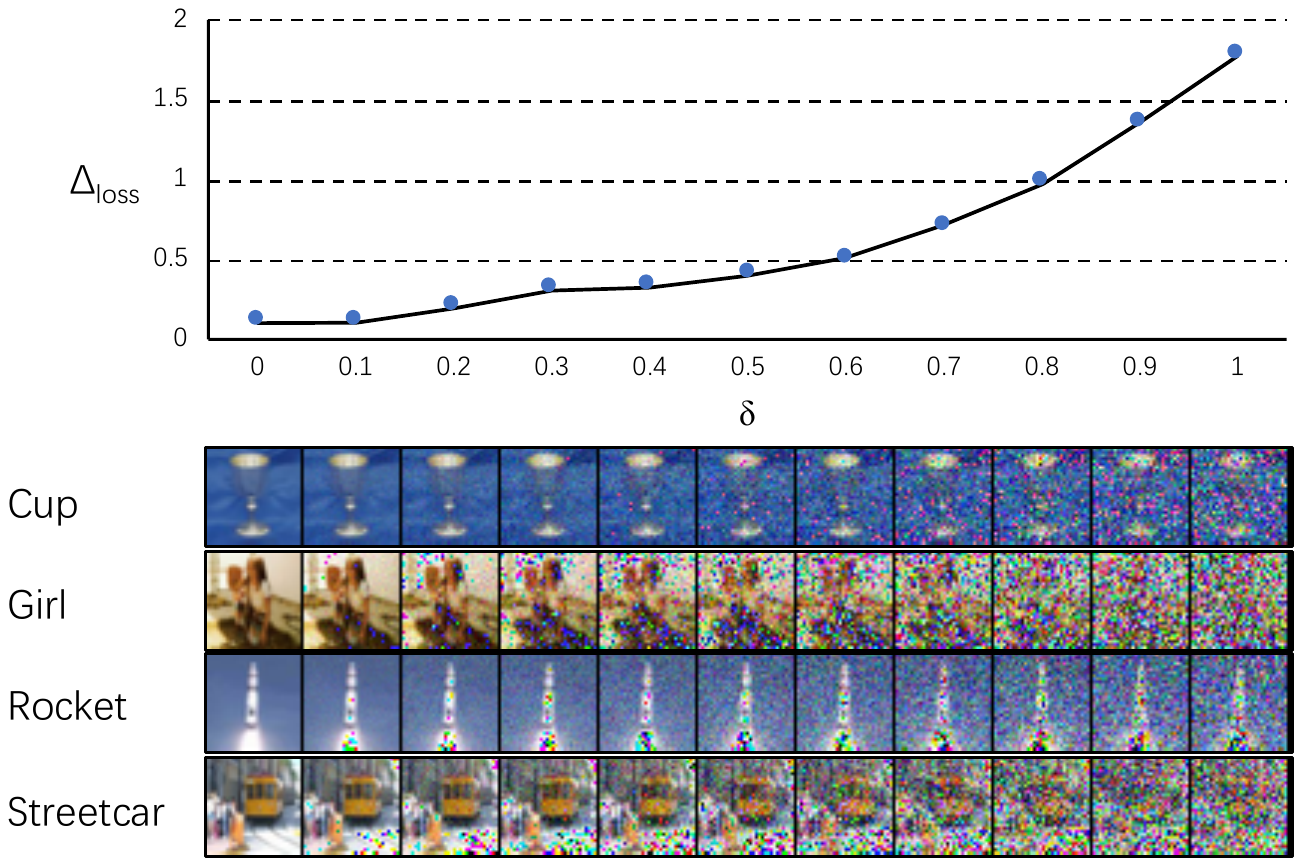}
 \caption{The curve of loss gap between KD loss and RCO loss along with Gaussian noise. $\delta$ represents the scale of Gaussian noise. $\Delta_{loss}$ represents the loss gap between KD loss to RCO loss. Larger $\Delta_{loss}$ means lower loss of RCO than KD. The bottom row shows part of noised images.
 }
 \label{fig:noise}
\end{figure}

\textbf{Visualization of Robustness to Noise.} Besides the visualization of optimization trajectory, we also observe that the new local minimum has better generalization capacity and is more robust to random noise in input space.
%In order to verify the robustness of the proposed method,
We consider bringing noise to the testing image. 
Firstly we calculate the standard deviation $\sigma_{in}$ for each image and set the $\delta$ ranging from 0.0 to 1.0 by step 0.1. 
The noise is sampled from $\mathcal{N}(0, \sigma^{2})$, where $\sigma=\sigma_{in} * \delta$.
We choose some noised images and show them at the bottom row of Figure \ref{fig:noise}.  The images are clear at first column, but as the $\delta$ increases, the images become illegible, especially for the last column. We ran this experiment on model trained both with KD and RCO and compared their loss. The loss gap from KD to RCO becomes more significant as the increasing of $\delta$, which suggests that model trained with RCO is more robust to noise than KD. We draw our result on top of Figure \ref{fig:noise}.

\section{Conclusion}
We have proposed a simple but effective and generally applicable method to boost the performance of small student network. By constructing an easy-to-hard sequence of learning target, small student network could achieve much higher performance compared with other knowledge transfer methods. 
Moreover, we offer two available strategies to construct the sequence of anchor points. For future work, we would like to explore the strategy to design the learning sequence automatically.

{\small
\bibliographystyle{ieee}
\bibliography{egbib}

\begin{thebibliography}{10}\itemsep=-1pt

\bibitem{Bengio2009Curriculum}
Y.~Bengio, J.~Louradour, R.~Collobert, and J.~Weston.
\newblock Curriculum learning.
\newblock In {\em International Conference on Machine Learning}, pages 41--48,
  2009.

\bibitem{Caruana2006Model}
R.~Caruana and A.~Niculescu-Mizil.
\newblock Model compression.
\newblock In {\em ACM SIGKDD International Conference on Knowledge Discovery
  and Data Mining}, pages 535--541, 2006.

\bibitem{deng2009imagenet}
J.~Deng, W.~Dong, R.~Socher, L.-J. Li, K.~Li, and L.~Fei-Fei.
\newblock Imagenet: A large-scale hierarchical image database.
\newblock In {\em Computer Vision and Pattern Recognition, 2009. CVPR 2009.
  IEEE Conference on}, pages 248--255. Ieee, 2009.

\bibitem{Deng2018ArcFace}
J.~Deng, J.~Guo, and S.~Zafeiriou.
\newblock Arcface: Additive angular margin loss for deep face recognition.
\newblock 2018.

\bibitem{goyal2017accurate}
P.~Goyal, P.~Doll{\'a}r, R.~Girshick, P.~Noordhuis, L.~Wesolowski, A.~Kyrola,
  A.~Tulloch, Y.~Jia, and K.~He.
\newblock Accurate, large minibatch sgd: training imagenet in 1 hour.
\newblock {\em arXiv preprint arXiv:1706.02677}, 2017.

\bibitem{guo2016ms}
Y.~Guo, L.~Zhang, Y.~Hu, X.~He, and J.~Gao.
\newblock Ms-celeb-1m: Challenge of recognizing one million celebrities in the
  real world.
\newblock {\em Electronic Imaging}, 2016(11):1--6, 2016.

\bibitem{han2015deep}
S.~Han, H.~Mao, and W.~J. Dally.
\newblock Deep compression: Compressing deep neural networks with pruning,
  trained quantization and huffman coding.
\newblock {\em arXiv preprint arXiv:1510.00149}, 2015.

\bibitem{han2015learning}
S.~Han, J.~Pool, J.~Tran, and W.~Dally.
\newblock Learning both weights and connections for efficient neural network.
\newblock In {\em Advances in neural information processing systems}, pages
  1135--1143, 2015.

\bibitem{He2016Deep}
K.~He, X.~Zhang, S.~Ren, and J.~Sun.
\newblock Deep residual learning for image recognition.
\newblock In {\em IEEE Conference on Computer Vision and Pattern Recognition},
  pages 770--778, 2016.

\bibitem{He2018AMC}
Y.~He, J.~Lin, Z.~Liu, H.~Wang, L.~J. Li, and S.~Han.
\newblock Amc: Automl for model compression and acceleration on mobile devices.
\newblock 2018.

\bibitem{hinton2014distilling}
G.~Hinton, O.~Vinyals, and J.~Dean.
\newblock Distilling the knowledge in a neural network.
\newblock 2014.

\bibitem{hinton2015distilling}
G.~Hinton, O.~Vinyals, and J.~Dean.
\newblock Distilling the knowledge in a neural network.
\newblock {\em arXiv preprint arXiv:1503.02531}, 2015.

\bibitem{howard2017mobilenets}
A.~G. Howard, M.~Zhu, B.~Chen, D.~Kalenichenko, W.~Wang, T.~Weyand,
  M.~Andreetto, and H.~Adam.
\newblock Mobilenets: Efficient convolutional neural networks for mobile vision
  applications.
\newblock {\em arXiv preprint arXiv:1704.04861}, 2017.

\bibitem{Hubara2016Quantized}
I.~Hubara, M.~Courbariaux, D.~Soudry, E.~Y. Ran, and Y.~Bengio.
\newblock Quantized neural networks: Training neural networks with low
  precision weights and activations.
\newblock {\em Journal of Machine Learning Research}, 18, 2016.

\bibitem{ioffe2015batch}
S.~Ioffe and C.~Szegedy.
\newblock Batch normalization: Accelerating deep network training by reducing
  internal covariate shift.
\newblock {\em arXiv preprint arXiv:1502.03167}, 2015.

\bibitem{krizhevsky2009learning}
A.~Krizhevsky and G.~Hinton.
\newblock Learning multiple layers of features from tiny images.
\newblock Technical report, Citeseer, 2009.

\bibitem{li2016pruning}
H.~Li, A.~Kadav, I.~Durdanovic, H.~Samet, and H.~P. Graf.
\newblock Pruning filters for efficient convnets.
\newblock {\em arXiv preprint arXiv:1608.08710}, 2016.

\bibitem{li2017visualizing}
H.~Li, Z.~Xu, G.~Taylor, and T.~Goldstein.
\newblock Visualizing the loss landscape of neural nets.
\newblock {\em arXiv preprint arXiv:1712.09913}, 2017.

\bibitem{li2017mimicking}
Q.~Li, S.~Jin, and J.~Yan.
\newblock Mimicking very efficient network for object detection.
\newblock In {\em 2017 IEEE Conference on Computer Vision and Pattern
  Recognition (CVPR)}, pages 7341--7349. IEEE, 2017.

\bibitem{liu2017learning}
Z.~Liu, J.~Li, Z.~Shen, G.~Huang, S.~Yan, and C.~Zhang.
\newblock Learning efficient convolutional networks through network slimming.
\newblock In {\em Computer Vision (ICCV), 2017 IEEE International Conference
  on}, pages 2755--2763. IEEE, 2017.

\bibitem{loshchilov1608sgdr}
I.~Loshchilov and F.~Hutter.
\newblock Sgdr: Stochastic gradient descent with warm restarts. 2016.
\newblock {\em arXiv preprint arXiv:1608.03983}.

\bibitem{Luo2016Face}
P.~Luo, Z.~Zhu, Z.~Liu, X.~Wang, X.~Tang, P.~Luo, Z.~Zhu, Z.~Liu, X.~Wang, and
  X.~Tang.
\newblock Face model compression by distilling knowledge from neurons.
\newblock In {\em AAAI Conference on Artificial Intelligence}, 2016.

\bibitem{mirzadeh2019improved}
S.-I. Mirzadeh, M.~Farajtabar, A.~Li, and H.~Ghasemzadeh.
\newblock Improved knowledge distillation via teacher assistant: Bridging the
  gap between student and teacher.
\newblock {\em arXiv preprint arXiv:1902.03393}, 2019.

\bibitem{Molchanov2016Pruning}
P.~Molchanov, S.~Tyree, T.~Karras, T.~Aila, and J.~Kautz.
\newblock Pruning convolutional neural networks for resource efficient
  inference.
\newblock 2016.

\bibitem{romero2014fitnets}
A.~Romero, N.~Ballas, S.~E. Kahou, A.~Chassang, C.~Gatta, and Y.~Bengio.
\newblock Fitnets: Hints for thin deep nets.
\newblock {\em arXiv preprint arXiv:1412.6550}, 2014.

\bibitem{sandler2018mobilenetv2}
M.~Sandler, A.~Howard, M.~Zhu, A.~Zhmoginov, and L.-C. Chen.
\newblock Mobilenetv2: Inverted residuals and linear bottlenecks.
\newblock In {\em Proceedings of the IEEE Conference on Computer Vision and
  Pattern Recognition}, pages 4510--4520, 2018.

\bibitem{Gregor2016Do}
G.~Urban, K.~J. Geras, S.~E. Kahou, O.~Aslan, S.~Wang, R.~Caruana, A.~Mohamed,
  M.~Philipose, and M.~Richardson.
\newblock Do deep convolutional nets really need to be deep and convolutional?
\newblock {\em Nature}, 521, 2016.

\bibitem{wang2018devil}
F.~Wang, L.~Chen, C.~Li, S.~Huang, Y.~Chen, C.~Qian, and C.~C. Loy.
\newblock The devil of face recognition is in the noise.
\newblock {\em arXiv preprint arXiv:1807.11649}, 2018.

\bibitem{Wu2016Quantized}
J.~Wu, L.~Cong, Y.~Wang, Q.~Hu, and J.~Cheng.
\newblock Quantized convolutional neural networks for mobile devices.
\newblock In {\em Computer Vision and Pattern Recognition}, pages 4820--4828,
  2016.

\bibitem{yim2017gift}
J.~Yim, D.~Joo, J.~Bae, and J.~Kim.
\newblock A gift from knowledge distillation: Fast optimization, network
  minimization and transfer learning.
\newblock In {\em The IEEE Conference on Computer Vision and Pattern
  Recognition (CVPR)}, volume~2, 2017.

\bibitem{Zagoruyko2016Paying}
S.~Zagoruyko and N.~Komodakis.
\newblock Paying more attention to attention: Improving the performance of
  convolutional neural networks via attention transfer.
\newblock 2016.

\bibitem{Zhang2017ShuffleNet}
X.~Zhang, X.~Zhou, M.~Lin, and J.~Sun.
\newblock Shufflenet: An extremely efficient convolutional neural network for
  mobile devices.
\newblock 2017.

\bibitem{zhang2017deep}
Y.~Zhang, T.~Xiang, T.~M. Hospedales, and H.~Lu.
\newblock Deep mutual learning.
\newblock {\em arXiv preprint arXiv:1706.00384}, 6, 2017.

\bibitem{zhou2017rocket}
G.~Zhou, Y.~Fan, R.~Cui, W.~Bian, X.~Zhu, and K.~Gai.
\newblock Rocket launching: A universal and efficient framework for training
  well-performing light net.
\newblock {\em stat}, 1050:16, 2017.

\end{thebibliography}
}

\end{document}